# A MULTI-METHOD IMPORTANCE AND PERFORMANCE–EFFICIENCY ANALYSIS OF TOPOLOGICAL METRICS FOR NATURAL VISIBILITY GRAPH-BASED CYBER-ATTACK DETECTION

**Ph.D., Ali Melih KANCA**
Department of Computer Engineering, Faculty of Computing and Informatics Sciences, Karabuk University, Karabuk, Türkiye
ORCID ID: https://orcid.org/0000-0003-3271-7463
melihkanca@karabuk.edu.tr

**Prof. İlker TÜRKER**
Department of Computer Engineering, Faculty of Computing and Informatics Sciences, Karabuk University, Karabuk, Türkiye
ORCID ID: https://orcid.org/0000-0001-7577-4658
iturker@karabuk.edu.tr

## ABSTRACT

Natural Visibility Graph (NVG)-based analysis enables network traffic to be characterized through multiple topological descriptors reflecting different structural properties of the resulting graphs. However, the relevance of these descriptors to cyber-attack classification may vary considerably, while extracting a large set of graph metrics can also increase computational cost. Determining which topological metrics are most informative and whether classification capability can be maintained using a reduced metric set is therefore important for improving both the interpretability and computational efficiency of NVG-based cyber-attack detection. This study systematically evaluates 21 NVG-derived topological metrics covering centrality, path length, clustering coefficient, community structure, graph diameter, and degree-based characteristics. Four complementary importance-analysis methods—SHAP, grouped Permutation Importance, Boruta, and Recursive Feature Elimination (RFE)—are employed, and their rankings are integrated through a Consensus Ranking strategy. Based on the resulting ranking, progressively reduced configurations comprising Full21, Top15, Top10, Top7, Top5, and Top3 metrics are evaluated in terms of classification performance and computational cost. Experiments on the CSE-CIC-IDS2018 dataset using a CNN classifier and stratified 5-fold cross-validation identify avg_clustering_coeff_median, avg_clustering_coeff_std, and avg_clustering_coeff_mean as the three highest-ranked metrics. Among the evaluated configurations, Top3 achieved the highest observed mean predictive performance, with 97.148% accuracy, 97.055% weighted F1-score, and an MCC of 0.9675, compared with 95.999% accuracy, 95.521% weighted F1-score, and an MCC of 0.9549 for Full21. Moreover, Top3 reduces total runtime from 14,961.39 s to 589.22 s, corresponding to a 96.06% reduction. These findings indicate that, under the evaluated experimental setting, importance-guided metric reduction can yield a compact NVG representation associated with higher observed mean predictive performance while substantially reducing computational cost.

## INTRODUCTION

Accurate identification of malicious network activities remains a fundamental requirement for protecting modern networked systems against diverse cyber threats. Machine learning (ML) and deep learning (DL) approaches have demonstrated strong capabilities for network intrusion detection; however, their effectiveness depends not only on the classification model but also on how network traffic is represented.

Graph-based representations provide an alternative to conventional flow-level analysis by capturing structural relationships among sequential traffic observations. In particular, Natural Visibility Graphs (NVGs) transform numerical sequences into graph structures while preserving their ordering and structural variation. These graphs can subsequently be characterized using topological descriptors such as centrality, clustering coefficient, path length, modularity, diameter, and degree statistics.

However, not all NVG-derived topological metrics necessarily contribute equally to cyber-attack classification. Moreover, extracting a large number of graph metrics can increase computational cost. Identifying the most informative metrics and determining whether compact metric subsets can preserve classification capability are therefore important for both interpretability and computational efficiency.

This study evaluates 21 NVG-derived topological metrics using four complementary approaches: SHAP, grouped Permutation Importance, Boruta, and Recursive Feature Elimination (RFE). Their rankings are integrated through a Consensus Ranking strategy. The resulting ranking is then evaluated using progressively reduced configurations—Full21, Top15, Top10, Top7, Top5, and Top3—to examine the effect of metric reduction on classification performance and computational cost.

The main contributions of this study are threefold: (i) a multi-method importance analysis of 21 NVG-derived topological metrics, (ii) a consensus-based evaluation of progressively reduced metric configurations, and (iii) a performance–efficiency analysis examining whether compact NVG representations can maintain predictive capability while reducing computational requirements.

## RELATED WORK

Machine learning and deep learning have been extensively investigated for network intrusion detection, with previous studies demonstrating their effectiveness in identifying complex malicious traffic patterns [1–3]. However, conventional approaches generally rely on flow-level or statistical traffic attributes, motivating the investigation of alternative representations capable of capturing structural relationships within network traffic.

Graph-based representations have consequently received increasing attention in cybersecurity. Previous studies have modeled hosts, communication entities, and network interactions as graph structures and applied graph-based machine learning or graph neural networks to intrusion and botnet detection. Chowdhury et al. [4] employed graph-based feature clustering for botnet detection, while Daya et al. [5] investigated graph-based machine learning for identifying bot activity. Graph-based intrusion detection has also been studied in Controller Area Networks [6], IoT environments [7,8], and network traffic analysis [9]. Furthermore, Shinan et al. [10] demonstrated the relevance of graph centrality measures for graph-based bot detection. Collectively, these studies indicate that structural graph characteristics can provide useful information for cybersecurity analysis.

Visibility Graph (VG) techniques provide another mechanism for obtaining graph-based representations by transforming sequential observations into complex networks. Lacasa et al. [11] introduced the visibility graph approach for mapping time-series observations into graph structures while preserving important characteristics of the original sequence. Subsequent research has demonstrated that graph-derived characteristics obtained from visibility-based representations can provide informative features for classification tasks. For example, visibility-graph-based feature extraction has been investigated for fault diagnosis [12], while graph-based feature extraction has also been explored for signal classification outside conventional time-frequency representations[13]. These findings support the use of graph topology as an alternative representation of sequential data.

In the context of network traffic analysis, the resulting graph structures can be characterized using topological descriptors such as centrality, clustering coefficient, degree statistics, shortest-path characteristics, modularity, and diameter. However, different topological metrics characterize different structural properties, and their contributions to a classification model may not be equivalent. Therefore, identifying which graph-derived descriptors provide the most relevant information is important not only for feature analysis but also for improving the interpretability of graph-based cyber-attack detection models.

Feature-importance and feature-selection techniques provide complementary mechanisms for evaluating such relevance. SHAP, introduced by Lundberg and Lee [14], provides a theoretically grounded framework for quantifying the contribution of individual features to model predictions. Boruta evaluates feature relevance by comparing original features with randomized shadow features [15], whereas Recursive Feature Elimination (RFE) progressively removes less informative features to establish their relative relevance [16]. Random Forest models can additionally serve as underlying estimators for model-based feature-selection procedures such as Boruta and RFE [17]. Because these techniques rely on different principles, the importance assigned to the same feature may vary considerably across methods.

Kanca and Türker [18] investigated traffic feature–topology combinations through statistical analyses of feature-wise NVG representations to characterize class-specific topological signatures of cyber-attacks. Their study focused on identifying structural differences among attack classes using a fixed set of ten graph-theoretic metrics.

Despite these developments, limited attention has been given to systematically comparing the importance of individual NVG-derived topological metric types for cyber-attack classification. Moreover, the extent to which importance-guided metric reduction can preserve predictive performance while reducing computational cost remains insufficiently investigated. In contrast to the class-specific structural analysis conducted by Kanca and Türker (2026), the present study focuses on the relative importance and computational utility of an expanded set of 21 topological metrics. To this end, SHAP, grouped Permutation Importance, Boruta, and RFE are integrated through a Consensus Ranking strategy, and progressively reduced ranked metric configurations are subsequently evaluated in terms of classification performance and computational efficiency.

## METHODOLOGY

### A. Dataset and Preprocessing

The experiments were conducted using the CSE-CIC-IDS2018 dataset, which contains benign traffic and multiple cyber-attack scenarios. Considering the computational requirements of NVG construction and topological metric extraction, a subset of 100,000 network-flow records covering 15 classes was constructed through class-wise sampling to obtain an approximately balanced record-level class distribution. No synthetic oversampling technique, including SMOTE, was applied.

Class labels were converted into numerical identifiers, and the selected traffic attributes were converted into numerical values. Samples containing missing, infinite, invalid, or non-convertible values were removed, while attributes unsuitable for numerical graph-based analysis were excluded. The resulting numerical traffic features were used as sequential observations for NVG construction.

**B. Sliding-Window-Based Frame Generation**

To capture local relationships among consecutive network-flow observations, the preprocessed data were segmented into overlapping frames using a sliding window of 100 observations with a step size of 50, corresponding to 50% overlap between consecutive frames.

Each frame was assigned a single class label using an attack-prioritized majority-voting strategy. Frames containing only benign observations were labeled as benign; otherwise, the most frequent non-benign class was assigned. In the case of a tie between non-benign classes, the first occurring non-benign class within the frame was selected. This procedure preserved attack information in frames containing both benign and malicious traffic.

**C. Natural Visibility Graph Construction**

Following frame generation, each numerical traffic feature within a frame was independently treated as a sequential signal and transformed into a Natural Visibility Graph (NVG). Thus, an individual NVG was constructed for each numerical traffic feature within each frame.

Given a numerical sequence $X = \{x_1, x_2, \ldots, x_n\}$, each observation $x_i$ is represented as a node in the NVG. Two observations $x_i$ and $x_j$, where $i < j$, are connected by an edge if every intermediate observation $x_k$, where $i < k < j$, satisfies the natural visibility criterion defined in Equation (1):

$$x_k < x_i + (x_j - x_i)\frac{k - i}{j - i} \quad (1)$$

This criterion connects two observations when the straight line between them is not intersected by any intermediate observation, thereby preserving the ordering and structural variation of the sequence. For metrics requiring a connected graph, the largest connected component was retained when necessary.

**D. Topological Metric Extraction**

A total of 21 topological metric types were extracted from each feature-specific NVG to characterize different structural properties. The evaluated metrics were grouped as follows:

**Centrality:** closeness_mean, betweenness_mean, degree_centrality_mean, eigenvector_mean.

**Path length:** avg_path_length_mean, avg_path_length_std, avg_path_length_median, avg_path_length_min, avg_path_length_max.

**Clustering coefficient:** avg_clustering_coeff_mean, avg_clustering_coeff_std, avg_clustering_coeff_median, avg_clustering_coeff_min, avg_clustering_coeff_max.

**Community/global structure:** modularity, diameter.

**Degree:** avg_degree_mean, avg_degree_std, avg_degree_median, avg_degree_min, avg_degree_max.

For each frame, these descriptors were calculated independently for every feature-specific NVG, and the resulting metric vectors were concatenated across the original traffic features to construct the graph-based representation used for classification.

### E. CNN-Based Classification

A multi-input Convolutional Neural Network (CNN) was employed to classify the NVG-based representations. Each original traffic feature was represented by an independent input branch containing its corresponding topological descriptors. Each branch consisted of a 1D convolutional layer with 32 filters (kernel size = 3), followed by batch normalization, dropout (0.2), and flattening. The branch outputs were concatenated and passed through a 128-neuron ReLU layer and a Softmax output layer for 15-class classification.

Model evaluation was performed using stratified 5-fold cross-validation. Within each fold, StandardScaler was fitted only on the training partition and then applied to both training and test data. The CNN was trained for 20 epochs with a batch size of 32 using Adam (learning rate = 0.001) and sparse categorical cross-entropy.

### F. Topological Metric Importance Analysis

The relative importance of the 21 NVG-derived topological metrics was evaluated using four complementary methods: SHAP, grouped Permutation Importance, Boruta, and Recursive Feature Elimination (RFE).

1) SHAP: GradientExplainer was applied to the trained CNN using up to 100 training samples as background data and 200 test samples for explanation. Absolute SHAP values were averaged across samples and original traffic-feature groups to obtain a single importance score for each metric type, with higher values indicating greater importance.

2) Permutation Importance: Each metric type was simultaneously permuted across all feature-specific NVG groups, and the resulting decrease in weighted F1-score was used as its importance score. The procedure was repeated five times per metric, with larger performance decreases indicating greater importance.

3) Boruta: A 21-dimensional metric-level representation was constructed by averaging each metric across the original traffic-feature groups. Boruta was applied using a Random Forest classifier with 300 trees and balanced class weights. Rankings and selection frequencies were aggregated across the cross-validation folds.

4) Recursive Feature Elimination: RFE was applied to the same 21-dimensional representation using a Random Forest classifier with 300 trees and balanced class weights. Metrics were iteratively eliminated according to their importance until a complete ranking was obtained.

The rankings produced by the four methods were subsequently integrated through the Consensus Ranking described in the following subsection.

### G. Consensus Ranking

Since the four feature importance and selection methods evaluate relevance from different perspectives, their individual rankings were integrated into a unified Consensus Ranking. For each topological metric $j$, the Consensus Score was calculated as the arithmetic mean of the ranks obtained from SHAP, grouped Permutation Importance, Boruta, and RFE, as defined in Equation (2):

$$CS_j = \frac{1}{4}\left(r_{j,SHAP} + r_{j,Permutation} + r_{j,Boruta} + r_{j,RFE}\right) \quad (2)$$

where $CS_j$ denotes the Consensus Score of topological metric j, and $r_{j,m}$ represents the rank assigned to metric j by feature importance method m. A lower Consensus Score indicates that a metric was consistently ranked as more important across the four methods. The 21 topological metrics were sorted in ascending order according to their Consensus Scores to obtain the final Consensus Ranking.

### H. Ranked Metric Subset Evaluation

Based on the final Consensus Ranking aggregated across the five cross-validation folds, six progressively reduced metric configurations were evaluated: Full21, Top15, Top10, Top7, Top5, and Top3. Each Top-K configuration retained the highest-ranked metric types and was subsequently evaluated using the same CNN architecture, training parameters, and stratified 5-fold cross-validation protocol. Classification performance was compared using Accuracy, Weighted F1-score, and MCC. Since the Consensus Ranking and the subsequent subset evaluation were derived from the same dataset and cross-validation framework, the Top-K experiments were treated as an internal comparative analysis of ranked metric configurations rather than as an independent validation of the selected subsets.

### I. Computational Cost Analysis

In addition to predictive performance, the computational cost of each ranked metric configuration was evaluated. For each configuration, the total experiment runtime included frame generation, NVG construction and extraction of the selected topological metrics, and the complete stratified 5-fold CNN evaluation. Runtime was measured separately for Full21, Top15, Top10, Top7, Top5, and Top3 using the same experimental pipeline. Importantly, only the topological metrics retained in each configuration were computed during metric extraction, rather than computing the complete set of 21 metrics and subsequently discarding unselected metrics. The relative runtime reduction of each reduced configuration was calculated with respect to the Full21 baseline.

$$RR_k = \frac{T_{Full21} - T_k}{T_{Full21}} \times 100$$

## EXPERIMENTAL RESULTS

### A. Baseline Classification Performance

The complete set of 21 topological metrics (Full21) was first evaluated as the baseline configuration using stratified 5-fold cross-validation. The model achieved an accuracy of **95.999 ± 2.256%**, a weighted F1-score of **95.521 ± 2.975%**, and an MCC of **0.9549 ± 0.0245**, demonstrating strong classification performance and providing the reference point for the subsequent importance and ranked-subset analyses.

### B. Individual Topological Metric Importance Analysis

The 21 NVG-derived topological metrics exhibited both agreements and differences across SHAP, grouped Permutation Importance, Boruta, and RFE. Several clustering-coefficient-based descriptors, particularly the median, standard deviation, and mean, consistently appeared among the higher-ranked metrics, indicating the relevance of local connectivity characteristics within the NVG representation.

Figure 1 compares the rankings produced by the four methods. The observed differences demonstrate that metric importance varies according to the evaluation mechanism, supporting the use of a multi-method approach rather than relying on a single ranking method.

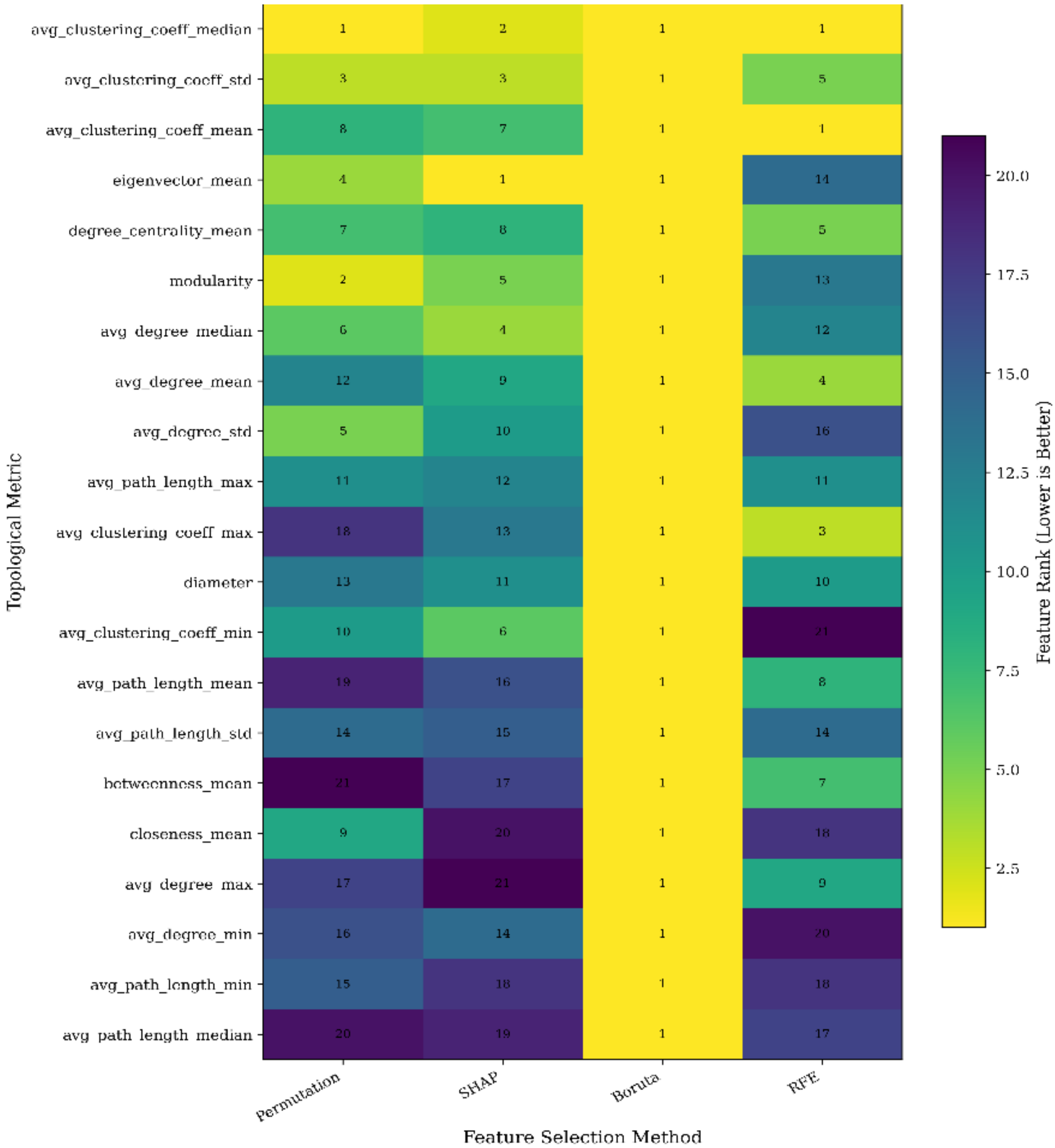

**Figure 1.** Comparison of topological metric rankings across the four importance methods.

## C. Consensus Ranking Results

The rankings obtained from SHAP, grouped Permutation Importance, Boruta, and RFE were integrated using the Consensus Score defined in Equation (2). Figure 2 presents the resulting

Consensus Ranking, while Table 1 reports the individual method ranks and Consensus Scores for all 21 NVG-derived topological metrics.

As shown in Figure 2 and Table 1, clustering-coefficient-based descriptors dominated the highest positions. avg_clustering_coeff_median ranked first with a Consensus Score of 1.25, followed by avg_clustering_coeff_std (3.00) and avg_clustering_coeff_mean (4.25). The next highest-ranked metrics were eigenvector_mean (5.00), degree_centrality_mean (5.25), and modularity (5.25). These results indicate that local connectivity characteristics represented by clustering-coefficient statistics were particularly informative under the evaluated setting.

Differences among the individual methods were also observed. For example, eigenvector_mean ranked first according to SHAP but 14th according to RFE, whereas modularity ranked second according to grouped Permutation Importance but 13th according to RFE. In contrast, Boruta selected all 21 metrics across the evaluated folds, resulting in an aggregated rank of 1 for every metric; thus, Boruta confirmed their relevance but did not discriminate their relative importance. These variations further support the use of a consensus-based ranking rather than relying on a single importance method.

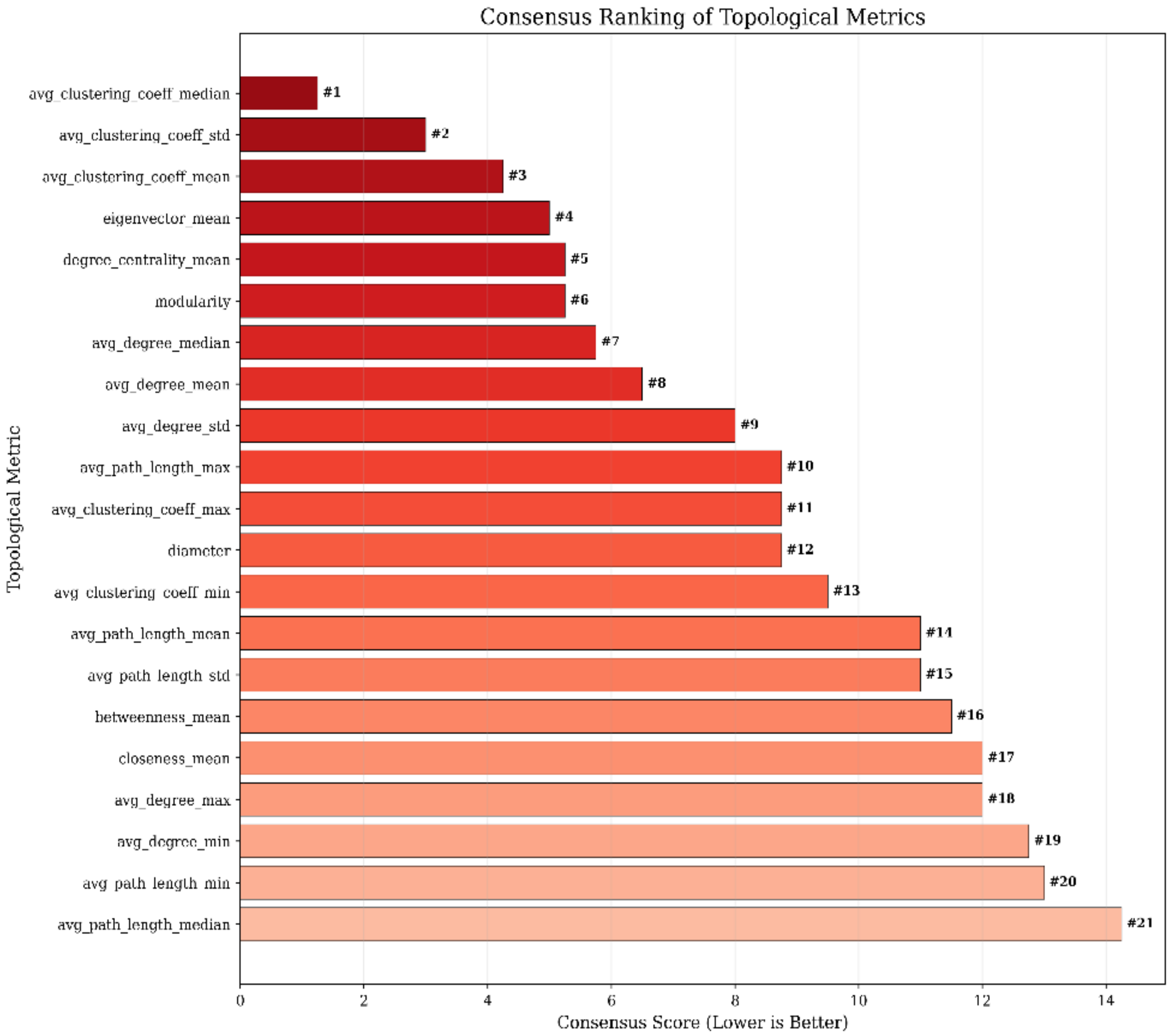

**Figure 2.** Consensus ranking of the 21 NVG-based topological metrics.

**Table 1.** Consensus ranking of the 21 NVG-derived topological metrics across the four importance methods.

| Final | Topological Metric | SHAP | Perm. | Boruta | RFE | CS |
|---|---|---|---|---|---|---|
| **1** | avg_clustering_coeff_median | 2 | 1 | 1 | 1 | 1.25 |
| **2** | avg_clustering_coeff_std | 3 | 3 | 1 | 5 | 3.00 |
| **3** | avg_clustering_coeff_mean | 7 | 8 | 1 | 1 | 4.25 |
| **4** | eigenvector_mean | 1 | 4 | 1 | 14 | 5.00 |
| **5** | degree_centrality_mean | 8 | 7 | 1 | 5 | 5.25 |
| **6** | modularity | 5 | 2 | 1 | 13 | 5.25 |
| **7** | avg_degree_median | 4 | 6 | 1 | 12 | 5.75 |
| **8** | avg_degree_mean | 9 | 12 | 1 | 4 | 6.50 |
| **9** | avg_degree_std | 10 | 5 | 1 | 16 | 8.00 |
| **10** | avg_path_length_max | 12 | 11 | 1 | 11 | 8.75 |
| **11** | avg_clustering_coeff_max | 13 | 18 | 1 | 3 | 8.75 |
| **12** | diameter | 11 | 13 | 1 | 10 | 8.75 |
| **13** | avg_clustering_coeff_min | 6 | 10 | 1 | 21 | 9.50 |
| **14** | avg_path_length_mean | 16 | 19 | 1 | 8 | 11.00 |
| **15** | avg_path_length_std | 15 | 14 | 1 | 14 | 11.00 |
| **16** | betweenness_mean | 17 | 21 | 1 | 7 | 11.50 |
| **17** | closeness_mean | 20 | 9 | 1 | 18 | 12.00 |
| **18** | avg_degree_max | 21 | 17 | 1 | 9 | 12.00 |
| **19** | avg_degree_min | 14 | 16 | 1 | 20 | 12.75 |
| **20** | avg_path_length_min | 18 | 15 | 1 | 18 | 13.00 |
| **21** | avg_path_length_median | 19 | 20 | 1 | 17 | **14.25** |

### D. Performance Evaluation of Ranked Metric Subsets

To evaluate whether the Consensus Ranking could support a more compact NVG-based representation, progressively reduced configurations (Top15, Top10, Top7, Top5, and Top3) were compared with the Full21 baseline. Table 2 summarizes their classification performance and computational cost.

As shown in Table 2, reducing the number of retained metrics did not necessarily degrade classification performance. Top10, Top7, and Top5 achieved weighted F1-scores of 96.419%, 96.323%, and 96.025%, respectively, all exceeding the Full21 baseline of 95.521%, while reducing total runtime by 28.23–47.95%. The Top15 configuration produced a slightly lower weighted F1-score (95.254%) while still reducing runtime by 21.15%.

**Table 2.** Classification performance and computational cost of the ranked metric configurations.

| Configuration | Accuracy (%) | Weighted F1 (%) | MCC | Total Runtime (s) | Reduction (%) |
|---|---|---|---|---|---|
| Full21 | 95.999 ± 2.256 | 95.521 ± 2.975 | 0.9549 ± 0.0245 | 14,961.39 | - |
| Top15 | 95.698 ± 1.373 | 95.254 ± 2.009 | 0.9515 ± 0.0148 | 11,797.47 | 21.15 |
| Top10 | 96.498 ± 0.588 | 96.419 ± 0.692 | 0.9602 ± 0.0066 | 10,738.08 | 28.23 |
| Top7 | 96.498 ± 0.968 | 96.323 ± 1.215 | 0.9603 ± 0.0107 | 9,480.09 | 36.64 |
| Top5 | 96.298 ± 1.291 | 96.025 ± 1.725 | 0.9581 ± 0.0140 | 7,787.95 | 47.95 |
| Top3 | 97.148 ± 0.823 | 97.055 ± 0.791 | 0.9675 ± 0.0093 | 589.22 | 96.06 |

Among the evaluated configurations, the highest observed mean performance was obtained with Top3, which retained only avg_clustering_coeff_median, avg_clustering_coeff_std, and avg_clustering_coeff_mean. Top3 achieved an accuracy of 97.148%, a weighted F1-score of 97.055%, and an MCC of 0.9675, while reducing total runtime from 14,961.39 s to 589.22 s, corresponding to a 96.06% reduction relative to Full21. These results describe the comparative behavior of the ranked configurations within the present experimental protocol.

The higher observed mean performance of the Top3 configuration may be associated with the removal of less informative or potentially redundant topological descriptors. Restricting the representation to the highest-ranked metrics may allow the CNN to focus on more discriminative structural information while reducing the influence of less informative or redundant descriptors. This reduction may also have a regularizing effect by decreasing the complexity of the input representation. However, these interpretations should be regarded as possible explanations for the observed behavior rather than as directly established causal mechanisms. Figure 3 illustrates the performance–efficiency relationship relative to the Full21 baseline.

The results show that the highest-ranked compact configurations can maintain or achieve higher observed mean predictive performance while substantially reducing computational cost under the evaluated experimental setting.

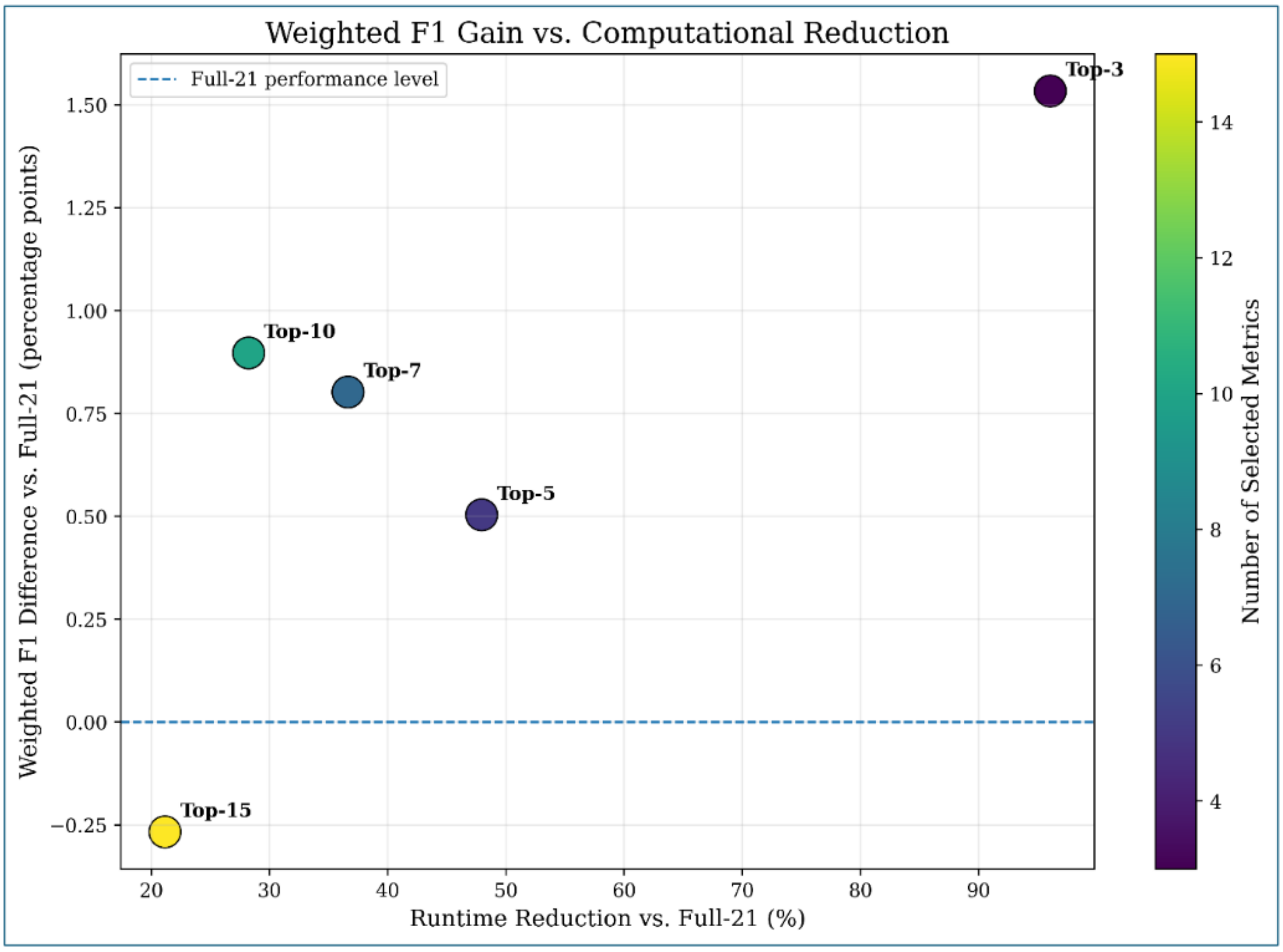

**Figure 3.** Weighted F1-score difference and runtime reduction of the ranked metric subsets relative to the Full21 baseline.

## CONCLUSION

This study investigated the relative importance and practical utility of 21 topological metrics derived from Natural Visibility Graph (NVG) representations for cyber-attack classification. Four complementary importance-analysis methods—SHAP, grouped Permutation Importance, Boruta, and Recursive Feature Elimination (RFE)—were employed to evaluate metric relevance from different perspectives, and their rankings were integrated through a Consensus Ranking strategy. The results demonstrated that the importance of NVG-derived descriptors is not uniformly distributed across topological metric types and that individual importance methods can produce substantially different rankings. These differences highlight the value of combining multiple importance mechanisms when assessing structural descriptors rather than relying on a single ranking criterion.

The Consensus Ranking identified avg_clustering_coeff_median, avg_clustering_coeff_std, and avg_clustering_coeff_mean as the three highest-ranked metrics, with Consensus Scores of 1.25, 3.00, and 4.25, respectively. Their concentration at the top of the ranking indicates that clustering-based descriptors of local connectivity constitute particularly informative components of the NVG representation under the evaluated experimental setting. Other structural descriptors, including eigenvector_mean, degree_centrality_mean, and modularity, also occupied high positions, suggesting that both local connectivity and broader graph-structural characteristics contribute to the representation. At the same time, the disagreements observed across SHAP, Permutation Importance, and RFE demonstrate that metric relevance is method-dependent, highlighting the value of a multi-method assessment. Boruta selected all 21 metrics across the evaluated folds and therefore did not provide discriminative ranking information among the metrics in this experimental setting. Consequently, its contribution to the Consensus Ranking was limited to confirming metric relevance rather than differentiating their relative importance.

The internal comparative experiments indicate that importance-guided metric reduction may provide predictive and computational benefits under the evaluated experimental setting. Reducing the representation from Full21 to Top10, Top7, and Top5 maintained or improved weighted F1-score while progressively reducing computational cost. Most notably, Top3, consisting only of avg_clustering_coeff_median, avg_clustering_coeff_std, and avg_clustering_coeff_mean, achieved the highest observed mean performance among the evaluated configurations, with an accuracy of 97.148%, weighted F1-score of 97.055%, and MCC of 0.9675. In comparison, Full21 achieved 95.999% accuracy, 95.521% weighted F1-score, and an MCC of 0.9549. Top3 also reduced total runtime from 14,961.39 s to 589.22 s, corresponding to a 96.06% reduction. These findings indicate that retaining a larger number of NVG-derived metrics does not necessarily yield superior classification performance and that a compact, importance-guided representation can offer a favorable performance–efficiency trade-off.

Overall, the study provides an interpretable framework for identifying structurally informative NVG-derived metrics and demonstrates the practical value of evaluating their predictive contribution together with computational cost. Nevertheless, the reported findings are specific to the evaluated dataset, CNN architecture, and experimental setting. Future work should therefore examine the robustness of the identified metric subsets across additional intrusion-detection datasets and alternative classification models. In addition, metric-level runtime profiling could be performed to determine how individual topological descriptors contribute to the overall computational cost, particularly given the substantial runtime differences observed among the reduced configurations.

The Consensus Ranking and the subsequent Top-K evaluation were conducted within the same five-fold cross-validation framework. Therefore, the reported Top-K results primarily reflect the comparative performance of the ranked metric subsets within the current experimental setting. Future studies could further assess the robustness and generalizability of the identified metric subsets using independent holdout data or nested cross-validation.